\documentclass{article}
\PassOptionsToPackage{numbers, compress}{natbib}

\usepackage[preprint]{neurips_2026}
\usepackage[utf8]{inputenc}
\usepackage[T1]{fontenc}
\usepackage{hyperref}
\usepackage{url}
\usepackage{booktabs}
\usepackage{amsfonts}
\usepackage{nicefrac}
\usepackage{microtype}
\usepackage{xcolor}
\usepackage{graphicx}
\usepackage{wrapfig}
\usepackage{amsmath}
\usepackage{multirow}
\usepackage{amssymb}
\usepackage{makecell}
\usepackage{cite}

\title{EndPrompt: Efficient Long-Context Extension via Terminal Anchoring}

\author{
  \makebox[\textwidth][c]{
    \setlength{\tabcolsep}{3pt}
    \begin{tabular}{ccccc}
      \textbf{Han Tian$^{1,2*}$} & \textbf{Luxuan Chen$^{3,2*}$} & \textbf{Xinran Chen$^2$} & \textbf{Rui Kong$^2$} & \textbf{Fang Wang$^2$} \\[0.3ex]
      \textbf{Jiamin Chen$^2$} & \textbf{Jinman Zhao$^2$} & \textbf{Yuchen Li$^{2\dagger}$} & \textbf{Jiashu Zhao$^2$} & \textbf{Shuaiqiang Wang$^2$} \\[0.3ex]
      \multicolumn{5}{c}{\textbf{Haoyi Xiong$^4$} \quad \textbf{Linghe Kong$^3$} \quad \textbf{Dawei Yin$^{2\dagger}$}}
    \end{tabular}
  } \\[0.25in]
  \makebox[\textwidth][c]{
    \begin{tabular}{c}
      $^1$Nankai University \quad $^2$Baidu Inc. \\
      $^3$Shanghai Jiao Tong University \quad $^4$Independent Researcher \\[0.3ex]
      \texttt{\small tianhan@mail.nankai.edu.cn, yuchenli1230@gmail.com, yindawei@acm.org}
    \end{tabular}
  }
}
\date{}

\begin{document}

\maketitle
\let\thefootnote\relax\footnotetext{$^*$Co-first authors with equal contributions.}
\let\thefootnote\relax\footnotetext{$^\dagger$Corresponding author.}



\begin{abstract}
Extending the context window of large language models typically requires training on sequences at the target length, incurring quadratic memory and computational costs that make long-context adaptation expensive and difficult to reproduce. We propose \textbf{EndPrompt}, a method that achieves effective context extension using only short training sequences. The core insight is that exposing a model to long-range relative positional distances does not require constructing full-length inputs: we preserve the original short context as an intact first segment and append a brief terminal prompt as a second segment, assigning it positional indices near the target context length. This two-segment construction introduces both local and long-range relative distances within a short physical sequence while maintaining the semantic continuity of the training text--a property absent in chunk-based simulation approaches that split contiguous context. We provide a theoretical analysis grounded in Rotary Position Embedding and the Bernstein inequality, showing that position interpolation induces a rigorous smoothness constraint over the attention function, with shared Transformer parameters further suppressing unstable extrapolation to unobserved intermediate distances. Applied to LLaMA-family models extending the context window from 8K to 64K, EndPrompt achieves an average RULER score of 76.03 and the highest average on LongBench, surpassing LCEG (72.24), LongLoRA (72.95), and full-length fine-tuning (69.23) while requiring substantially less computation. These results demonstrate that long-context generalization can be induced from sparse positional supervision, challenging the prevailing assumption that dense long-sequence training is necessary for reliable context-window extension. The code is available at \url{https://github.com/clx1415926/EndPrompt}.

\end{abstract}

\section{Introduction}

Large language models are the foundation of modern natural language processing and a central interface for complex reasoning. However, their reliable maximum context length constrains utility. Applications such as long-document question answering~\citep{kovcisky2018narrativeqa}, repository-level code understanding~\citep{jimenez2024swe, guo2024deepseek}, legal and scientific document analysis~\citep{chalkidis2022lexglue}, personalized assistants, and extensive evidence retrieval require reasoning over inputs exceeding pretraining context windows. Here, longer contexts are effective only if models preserve local coherence and establish reliable interactions between distant tokens. Consequently, extending pretrained context windows is a central problem in language model adaptation.

Continuing to train models on long sequences at the target context length is a straightforward but costly solution. Collecting high-quality long-form corpora is difficult, and training increases memory consumption and runtime due to the quadratic scaling of attention computation~\citep{tay2022efficient}. While efficient implementations like FlashAttention~\citep{dao2022flashattention} and distributed systems mitigate this burden, full-length fine-tuning remains expensive. To reduce costs, recent methods explore position interpolation~\citep{chen2023extending, peng2024yarn, chen2024clex, ding2024longrope}, sparse attention~\citep{child2019generating}, sliding-window attention~\citep{beltagy2020longformer, xiao2024efficient, han2024lm}, low-rank adaptation~\citep{hu2021lora}, and simulated long-context training~\citep{zhu2024pose, bai2024longalign}. However, these approaches often introduce limitations, such as requiring substantial long-sequence training, altering the structure of attention, or chunking text in ways that damage semantic continuity. Thus, it remains unclear whether context extension requires dense supervision at the target length or if a sparse set of structured positional signals is sufficient.

This paper investigates whether models must observe full-length sequences to acquire long-context capabilities. We explore this question through the lens of positional generalization. In models utilizing Rotary Position Embedding~\citep{su2024roformer}, attention scores depend on both token content and relative positional distance. While existing methods assume reliable extrapolation requires exposure to dense relative distances during training, we demonstrate that effective long-context behavior emerges from sparse training signals. Specifically, models trained on short sequences can receive supervision for long-range positions if the examples preserve semantic coherence and provide stable anchors for distant positions. This finding reframes context adaptation as the design of informative positional supervision rather than merely increasing physical sequence length.

We propose \textbf{EndPrompt}, an efficient context-extension method utilizing positional index manipulation and an appended end prompt. We retain the original short context as the first segment and append a terminal prompt as the second. The first segment receives local positional indices, and the second receives indices near the target maximum context length. This configuration generates both local and long-range relative distances within a short physical sequence. Unlike chunk-based methods~\citep{zhu2024pose}, our approach avoids splitting contiguous text, thereby preserving semantic integrity while exposing models to long-distance positional patterns. The end prompt functions as a stable terminal anchor. Experiments indicate robustness across various prompt formulations, suggesting efficacy derives from the terminal cue's structural position rather than memorizing specific tokens. This design exposes long-range positional interactions without compromising the quality of the short-context training signal.

We analyze how sparse supervision supports long-context generalization. Under Rotary Position Embedding~\citep{su2024roformer}, attention scores act as a sum of sinusoidal components over relative distances, featuring content-dependent amplitudes and phases. Position interpolation~\citep{chen2023extending} reduces effective positional frequencies, constraining the attention score's variation rate and curvature across the distance dimension. This reduction induces a smoothness bias over unobserved intermediate distances. Furthermore, Transformers do not learn independent parameters for each distance; the same query and key projections support local and long-distance behavior. Consequently, sparse long-range supervision, combined with local supervision, constrains the shared parameter space and minimizes unstable behavior in unobserved regions. This theoretical perspective explains how short-sequence training produces stable long-context capabilities.

We evaluate \textbf{EndPrompt (ET)} on LLaMA-family models~\citep{touvron2023llama, grattafiori2024llama}, extending the context window from 8K to 64K. RULER~\citep{hsieh2024ruler} and LongBench~\citep{bai2024longbench} experiments show our method achieves competitive or superior performance compared to baselines such as LCEG~\citep{lu2024controlled}, LongLoRA~\citep{chen2024longlora}, and full-length fine-tuning. Specifically, our method achieves an average RULER score of 76.03, outperforming LCEG (72.24), LongLoRA (72.95), and full-length fine-tuning (69.23). On LongBench, our method secures the highest average score and demonstrates strong performance across tasks including question answering, summarization, few-shot learning, and code completion. Ablation studies validate the effects of end-prompt design, base model choice, extension length, and training-token quantity. These results confirm that reliable long-context adaptation emerges from structured sparse positional supervision without full-length training sequences.

The main contributions of this work are summarized as follows:
\begin{itemize}
    \item We propose \textbf{EndPrompt (ET)}, an efficient context-extension method utilizing short training sequences, positional index manipulation, and an appended end prompt to simulate long-range positional supervision.
    \item We demonstrate that preserving the original context as an undivided segment maintains semantic continuity, while the end prompt provides a stable terminal anchor to create long-distance relationships without disrupting the signal for next-token prediction.
    \item We analyze the method through Rotary Position Embedding and position interpolation, explaining how smooth positional variation and shared parameters of the Transformer support generalization over unobserved intermediate distances.
    \item We demonstrate strong empirical performance on RULER and LongBench, where our approach outperforms representative baselines while avoiding full-length training sequences.
\end{itemize}
\section{Preliminary}
\label{sec:preliminary}

This section reviews the positional mechanisms of the proposed method: RoPE~\citep{su2024roformer} and PI~\citep{chen2023extending}. 

For a given attention head, let $\mathbf{q}_m,\mathbf{k}_n \in \mathbb{R}^{D}$ denote the query and key vectors at positions $m$ and $n$. RoPE divides the dimensions into complex subspaces. In the $j$-th subspace, with assigned positional indices $p_m$ and $p_n$, RoPE applies a position-dependent phase rotation to the content components $q_{m,j}$ and $k_{n,j}$. This yields $\hat{q}_{m,j} = q_{m,j} e^{i p_m \theta_j}$ and $\hat{k}_{n,j} = k_{n,j} e^{i p_n \theta_j}$, where $\theta_j$ is the fixed angular frequency. The unnormalized attention score contribution from this subspace depends on the assigned relative distance $d=p_m-p_n$. By expressing the content term $q_{m,j}\bar{k}_{n,j}$ in polar form with amplitude $a_j(\mathbf{x}_m,\mathbf{x}_n;\Theta)$ and phase offset $\phi_j(\mathbf{x}_m,\mathbf{x}_n;\Theta)$, the total RoPE attention score becomes a finite trigonometric polynomial:
\begin{equation}
    S_{\mathrm{RoPE}}(d \mid \mathbf{x}_m,\mathbf{x}_n;\Theta)
    =
    \sum_{j=0}^{D/2-1}
    a_j(\mathbf{x}_m,\mathbf{x}_n;\Theta)
    \cos\left(d\theta_j+\phi_j(\mathbf{x}_m,\mathbf{x}_n;\Theta)\right).
    \label{eq:rope_score}
\end{equation}
Because the distance variable is exclusively embedded within the sinusoidal phase, modulating the positional indices enables attention over broader assigned relative distances without modifying the physical sequence length.

To adapt RoPE for extended contexts, PI rescales the positional indices by a target scale factor $s>1$, mapping $p$ to $p/s$. This operation effectively reduces the angular frequency to $\theta_j/s$, modifying the overall attention score to:
\begin{equation}
    S_{\mathrm{PI}}(d \mid \mathbf{x}_m,\mathbf{x}_n;\Theta)
    =
    \sum_{j=0}^{D/2-1}
    a_j(\mathbf{x}_m,\mathbf{x}_n;\Theta)
    \cos\left(\frac{d\theta_j}{s}+\phi_j(\mathbf{x}_m,\mathbf{x}_n;\Theta)\right).
    \label{eq:pi_score}
\end{equation}
Compared to Equation~\ref{eq:rope_score}, this rescaling lowers the maximum rate of change along the distance dimension. Because $S_{\mathrm{PI}}$ is a finite trigonometric polynomial, the maximum effective frequency $\theta_0$ strictly bounds the first-order variation and second-order curvature of the function:
\begin{equation}
    \sup_{d}\left|\frac{\partial S_{\mathrm{PI}}}{\partial d}\right|
    \le
    \frac{\theta_0}{s}
    \sup_{d}\left|S_{\mathrm{PI}}(d)\right|,
    \quad
    \sup_{d}\left|\frac{\partial^2 S_{\mathrm{PI}}}{\partial d^2}\right|
    \le
    \left(\frac{\theta_0}{s}\right)^2
    \sup_{d}\left|S_{\mathrm{PI}}(d)\right|.
    \label{eq:pi_smoothness}
\end{equation}
Rather than guaranteeing perfect reconstruction for unseen distances, these bounds indicate that PI provides a smoothness bias by suppressing high-frequency positional variations. The proposed method utilizes this smoothness, combined with targeted long-distance supervision, to stabilize attention scores across distances unobserved during training.
\section{Method}
\label{sec:method}

\begin{figure}[!htbp]
\centering
\includegraphics[width=.99\linewidth]{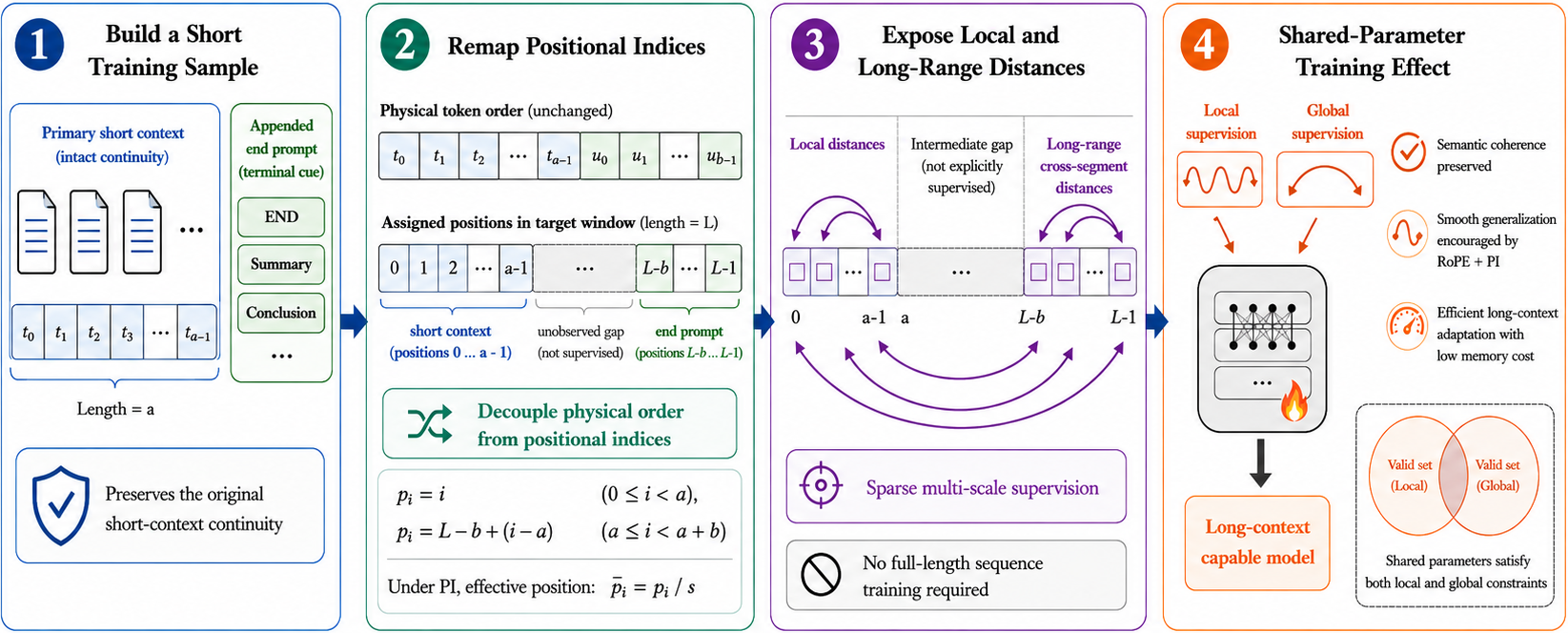}
\caption{Overview of the proposed method.}
\label{fig:overview}
\end{figure}

\subsection{Overview}
\label{subsec:method_overview}

As illustrated in Figure \ref{fig:overview}, the proposed method aims to achieve efficient long-context adaptation without the high memory and computational costs associated with full-length sequence training. This objective is realized through two coupled components. First, positional index manipulation decouples the physical token order from the assigned positional indices, which creates sparse long-distance supervision while maintaining local attention. Second, an appended end prompt acts as a terminal anchor near the boundary of the target context window. This design preserves the semantic integrity of the original short context. Together, these components enable the model to acquire long-range positional capabilities from short training samples under the frameworks of RoPE and PI.

\subsection{Positional Index Manipulation}
\label{subsec:positional_index_manipulation}

Let $L$ denote the target context length. Given a short context sequence $\mathbf{x}=(x_0,x_1,\ldots,x_{a-1})$ of length $a$, an end prompt $\mathbf{e}=(e_0,e_1,\ldots,e_{b-1})$ of length $b$ is appended to form the physical training sequence:
\begin{equation}
    \mathbf{y}
    =
    (x_0,x_1,\ldots,x_{a-1},e_0,e_1,\ldots,e_{b-1}),
    \quad
    |\mathbf{y}|=a+b.
    \label{eq:augmented_sequence}
\end{equation}
While the physical length is $a+b$, the assigned positional indices span both ends of the target context window via the mapping:
\begin{equation}
    p_{\ell}
    =
    \begin{cases}
    \ell, & 0 \le \ell < a, \\
    L-b+(\ell-a), & a \le \ell < a+b.
    \end{cases}
    \label{eq:position_mapping}
\end{equation}
Consequently, the short context and the end prompt are assigned to the intervals $[0,a-1]_{\mathbb{Z}}$ and $[L-b,L-1]_{\mathbb{Z}}$, respectively. With PI, the effective positional index becomes
\begin{equation}
    \bar{p}_{\ell}=\frac{p_{\ell}}{s},
    \label{eq:interpolated_position_mapping}
\end{equation}
for an interpolation scale factor $s>1$. Thus, the attention score evaluates the assigned relative distance $p_{\ell}-p_r$ instead of the physical relative distance $\ell-r$:
\begin{equation}
    S_{\ell r}
    =
    \sum_{j=0}^{D/2-1}
    a_{j,\ell r}(\Theta)
    \cos\left(
    \frac{(p_{\ell}-p_r)\theta_j}{s}
    +
    \phi_{j,\ell r}(\Theta)
    \right),
    \quad r\le \ell.
    \label{eq:method_attention_score}
\end{equation}
Equation~\ref{eq:method_attention_score} links the proposed position mapping with the PI score in Equation~\ref{eq:pi_score}, allowing the attention mechanism to receive positional phases corresponding to the long-context range despite a short physical sequence.

Under causal attention, the observed set of assigned relative distances comprises local intervals from the original context and the end prompt, alongside a long-range interval between them:
\begin{equation}
    D_{\mathrm{obs}}
    =
    [0,a-1]_{\mathbb{Z}}
    \cup
    [0,b-1]_{\mathbb{Z}}
    \cup
    [L-a-b+1,L-1]_{\mathbb{Z}}.
    \label{eq:observed_distance_set}
\end{equation}
Assuming $L-a-b\ge \max(a,b)$, the unobserved intermediate region is
\begin{equation}
    D_{\mathrm{gap}}
    =
    [\max(a,b),L-a-b]_{\mathbb{Z}}.
    \label{eq:gap_distance_set}
\end{equation}
Rather than explicitly supervising all distances in $D_{\mathrm{gap}}$, the model is trained on local and selected long distances. This mechanism relies on the smooth spectral structure of RoPE and PI to constrain behavior over the gap region, thereby providing sparse but multi-scale supervision for long-context adaptation.

\subsection{End Prompt as the Terminal Segment}
\label{subsec:end_prompt}

Splitting the original context to create long relative distances disrupts semantic continuity, as syntactic dependencies and local discourse relations rely on the original token order. Such splitting can remove essential local evidence for next-token prediction, degrading the quality of the supervision.

To circumvent this, the proposed method retains the intact original context and appends an end prompt as the terminal segment, assigned to the interval $[L-b,L-1]_{\mathbb{Z}}$ via Equation~\ref{eq:position_mapping}. This preserves local dependencies while establishing cross-segment distances approaching the target context length. The end prompt serves strictly as an explicit terminal cue rather than a semantic continuation.

Formally, the end prompt is sampled from a set of short terminal cues:
\begin{equation}
    \mathbf{e}\sim \mathcal{E},
    \label{eq:end_prompt_sampling}
\end{equation}
where $\mathcal{E}$ denotes the cue set. A unique prompt string is unnecessary; the critical factor is structural placement near the end of the assigned context window. Provided the prompt offers a stable terminal cue without conflicting semantics, various formulations can induce the requisite long-distance interactions, thereby mitigating the risk of prompt memorization and enhancing robustness.

\subsection{Training Objective}
\label{subsec:training_objective}

The proposed method integrates into standard autoregressive fine-tuning. Given the augmented sequence (Equation~\ref{eq:augmented_sequence}) and assigned positions (Equation~\ref{eq:position_mapping}), the training objective is
\begin{equation}
    \mathcal{L}(\Theta)
    =
    -
    \sum_{\ell=0}^{a+b-2}
    w_{\ell}
    \log
    P_{\Theta}
    \left(
    y_{\ell+1}
    \mid
    y_{\le \ell};
    p_{\le \ell}
    \right),
    \label{eq:training_objective}
\end{equation}
where $w_{\ell}\ge 0$ denotes an optional loss weight and $p_{\le \ell}$ determines the attention phases. In practice, prompt-token losses are assigned a smaller but nonzero weight. This design reduces excessive reliance on prompt-token prediction while preserving the loss signal on terminal tokens, whose causal attention can attend to the original context over large assigned positional distances.

This optimization imposes both local and global constraints on shared Transformer parameters, preventing the model from learning independent parameters for each distance. Local constraints emerge from predictions within the original context and, for $b$, within the end prompt, whereas global constraints arise from the nonzero terminal-token losses, through which terminal-segment states attend to the original segment across large assigned distances.

Expressing these constraints through feasible parameter sets, the admissible region under purely local supervision is
\begin{equation}
    \Theta_{\mathrm{local}}
    =
    \left\{
    \Theta:
    \mathcal{L}_{\mathrm{local}}(\Theta)
    \le
    \varepsilon_{\mathrm{local}}
    \right\}.
    \label{eq:local_feasible_set}
\end{equation}
With terminal long-distance supervision, the region becomes
\begin{equation}
    \Theta_{\mathrm{valid}}
    =
    \left\{
    \Theta:
    \mathcal{L}_{\mathrm{local}}(\Theta)
    \le
    \varepsilon_{\mathrm{local}}
    \right\}
    \cap
    \left\{
    \Theta:
    \mathcal{L}_{\mathrm{global}}(\Theta)
    \le
    \varepsilon_{\mathrm{global}}
    \right\}.
    \label{eq:valid_feasible_set}
\end{equation}
This reduction in the feasible region eliminates parameter configurations that fail to generalize to long distances, acting as an implicit regularizer over the attention function (Equation~\ref{eq:method_attention_score}).

\subsection{Connection to Smooth Long-Context Adaptation}
\label{subsec:smooth_long_context_adaptation}

The effectiveness of the proposed method originates from the synergy between sparse distance exposure and the spectral properties of RoPE and PI. Because the manipulated positional indices alter only the phase term in Equation~\ref{eq:method_attention_score}, the content-dependent amplitudes and phases remain governed by shared parameters. Consequently, the long-distance training signals act as a global constraint that directly regularizes the functions dictating local behavior. Furthermore, as indicated by Equation~\ref{eq:pi_smoothness}, PI reduces the effective angular frequency, which bounds the rate of positional variation and prevents unstable high-frequency oscillations within the unobserved gap region. In essence, the proposed method facilitates a constrained smooth extrapolation. The local and terminal supervisions anchor the short-range and long-range behaviors, PI suppresses excessive positional curvature, and the shared parameters unify these multi-scale constraints to achieve efficient long-context adaptation.

\section{Experiments}
\label{sec:experiments}


\subsection{Experimental Setup}
\label{subsec:setup}

The proposed method is evaluated on the architectures of LLaMA-2 7B \citep{touvron2023llama} and LLaMA-3 8B \citep{grattafiori2024llama}. The default configuration utilizes a corpus of one billion tokens to extend the context window from 8K to 64K. LongBench \citep{bai2024longbench} and RULER \citep{hsieh2024ruler} are utilized to evaluate the capabilities of the models in processing extended contexts. Furthermore, standard benchmarks, including GSM8K \citep{cobbe2021training}, HumanEval \citep{chen2021evaluating}, MMLU \citep{hendrycks2020measuring}, and HellaSwag \citep{zellers2019hellaswag}, are employed to assess the capabilities for short-text understanding. A comprehensive description of the evaluation tasks and the specific datasets is provided in Appendix \ref{subsec:tasks}.

\paragraph{Baselines}
We compare the proposed approach against four strong baselines: Positional Skip-Embedding \citep{zhu2024pose}, LCEG \citep{lu2024controlled}, LongLoRA \citep{chen2024longlora}, and full-length fine-tuning. Positional Skip-Embedding extends the context window by chunking inputs and manipulating position indices within a fixed window. LCEG provides a standardized protocol for evaluating the generalization of long contexts. LongLoRA accelerates the extension process using shifted sparse attention to minimize computational costs while retaining the original architectures. Finally, full-length fine-tuning trains the models directly on the target context length, serving as a resource-intensive standard for comparison.

\begin{table}[!ht]
  \caption{Results on the RULER synthetic benchmark (4K-64K).}
  \label{tab:ruler_results}
  \centering
  \small
  \setlength{\tabcolsep}{13pt} 
  \begin{tabular}{l | c c c c}
    \toprule
    \multirow{2}{*}{\textbf{Task}} & \multicolumn{4}{c}{\textbf{Method}} \\
    \cmidrule(lr){2-5}
    & \textbf{ET} & \textbf{LCEG} & \textbf{LongLoRA} & \textbf{Full FT} \\
    \midrule
    Niah\_S1 & 100.00 & 100.00 & 100.00 & 97.56 \\
    Niah\_S2 & 91.28  & 99.28  & 99.44  & 98.12 \\
    Niah\_S3 & 92.92  & 79.68  & 86.20  & 72.72 \\
    Niah\_M1 & 90.20  & 96.12  & 97.36  & 94.52 \\
    Niah\_M2 & 85.48  & 76.48  & 77.96  & 90.24 \\
    Niah\_M3 & 62.92  & 45.72  & 51.92  & 56.20 \\
    Niah\_MV & 81.67  & 77.81  & 81.56  & 62.34 \\
    Niah\_MQ & 82.06  & 83.88  & 83.79  & 62.56 \\
    Vt       & 82.00  & 68.18  & 65.70  & 68.56 \\
    Cwe      & 42.82  & 53.14  & 50.97  & 38.32 \\
    Fwe      & 83.53  & 63.01  & 58.17  & 58.10 \\
    Qa\_1    & 51.88  & 53.04  & 52.40  & 55.68 \\
    Qa\_2    & 41.60  & 42.84  & 42.92  & 45.04 \\
    \midrule
    Avg.     & \textbf{76.03}  & 72.24  & 72.95  & 69.23 \\
    \bottomrule
  \end{tabular}
\end{table}
\begin{table}[!ht]
  \caption{Results on various downstream tasks.}
  \label{tab:results_part1}
  \renewcommand{\arraystretch}{1.3}
\setlength{\tabcolsep}{4pt}
\small
\begin{tabular}{l ccccccc}
\toprule
\textbf{Method} & \makecell{\textbf{Single-Doc} \\ \textbf{QA}} & \makecell{\textbf{Multi-Doc} \\ \textbf{QA}} & \makecell{\textbf{Summari-} \\ \textbf{zation}} & \makecell{\textbf{Few-Shot} \\ \textbf{Learning}} & \makecell{\textbf{Synthetic} \\ \textbf{Task}} & \makecell{\textbf{Code Com-} \\ \textbf{pletion}} & \makecell{\textbf{Avg.} \\ \textbf{Score}} \\
\midrule
\midrule

ET & \textbf{\underline{32.03}} & \textbf{\underline{30.81}} & \textbf{\underline{26.04}} & \textbf{\underline{68.04}} & 4.54 & \textbf{\underline{66.48}} & \textbf{\underline{38.30}} \\
LCEG & 27.86 & 28.51 & 23.88 & 61.81 & 5.31 & 46.86 & 36.61 \\
LongLoRA & 26.86 & 27.51 & 22.88 & 60.81 & 4.31 & 45.86 & 36.84 \\
Full FT & 28.86 & 29.51 & 24.88 & 62.81 & \textbf{\underline{6.31}} & 47.86 & 35.63 \\ 
\midrule
\bottomrule
\end{tabular}
\end{table}

\subsection{Main Results on Long-Context Benchmarks}
\label{subsec:main_results}
Table \ref{tab:ruler_results} and Table \ref{tab:results_part1} present the performance of the models on RULER and LongBench. In the main comparison, models are trained on a one-billion-token corpus to extend the context window from 8K to 64K. The results demonstrate the ability of the proposed method to achieve superior performance in long-document understanding and retrieval compared to full-length fine-tuning and parameter-efficient baselines. Furthermore, we evaluate the training efficacy of the proposed method in terms of memory footprint and time consumption. The results indicate that our method effectively overcomes the traditional space-time trade-off, achieving significant reductions in memory utilization while simultaneously accelerating training speed compared to baseline methodologies. Detailed results and comprehensive analysis can be found in Appendix \ref{subsec:training_efficacy}.

\paragraph{Superior Overall Performance across Benchmarks} 
ET consistently achieves the highest average performance across both frameworks. On RULER, ET reaches an average score of 76.03, outperforming LongLoRA (72.95) and LCEG (72.24). On LongBench, the standard ET secures an average score of 38.30, exceeding full-length fine-tuning (35.63). This consistent gap highlights the effectiveness of ET across varied context lengths and tasks.

\paragraph{Resilience Divergence in Complex Information Retrieval} 
While the single-needle retrieval tasks (e.g., Niah\_S1) saturate at perfect scores of 100.00 for ET, LCEG, and LongLoRA, the performance diverges significantly as the complexity increases. ET demonstrates substantial robustness in demanding settings, outperforming LongLoRA on Vt (82.00 vs. 65.70) and Fwe (83.53 vs. 58.17). In multi-needle scenarios (Niah\_MV and Niah\_MQ), ET maintains strong scores of 81.67 and 82.06, respectively. These results indicate that the proposed method extends the context window without degrading the capacity to extract deeply embedded information.

\paragraph{Profound Advantages in Practical Downstream Applications} 
ET exhibits significant advantages in practical downstream applications. It achieves the highest scores across all realistic text processing domains, notably in Code Completion (66.48), significantly exceeding LCEG (46.86) and LongLoRA (45.86). A similar margin is observed in Few-Shot Learning, where ET scores 68.04 compared to a baseline average of approximately 61. By leading in Single-Doc QA, Multi-Doc QA, and Summarization, ET demonstrates that the architectural enhancements effectively translate to real-world semantic reasoning.



\begin{table}[!ht]
  \caption{Ablation study on the RULER synthetic benchmark across different configurations.}
  \label{tab:ruler_grouped_results}
  \centering
  \small
  \renewcommand{\arraystretch}{1.05}
  \resizebox{\textwidth}{!}{
  \begin{tabular}{l cccccccc c}
    \toprule
    \multirow{2}{*}{\textbf{Method}} &
    \multicolumn{8}{c}{\textbf{Task}} &
    \multirow{2}{*}{\textbf{Avg.}} \\
    \cmidrule(lr){2-9}
    & \textbf{Niah\_Single} & \textbf{Niah\_Multikey} & \textbf{Niah\_Multivalue} & \textbf{Niah\_Multiquery}
    & \textbf{Vt} & \textbf{Cwe} & \textbf{Fwe} & \textbf{Qa} & \\
    \midrule

    \multicolumn{10}{c}{\textit{Main}} \\
    \midrule
    ET & 94.73 & 79.53 & 81.67 & 82.06 & 82.00 & 42.82 & 83.53 & 46.74 & 76.03 \\
    \midrule

    \multicolumn{10}{c}{\textit{Different Base Models}} \\
    \midrule
    Mistral-7B-v0.3 & 98.75 & 70.92 & 62.07 & 87.36 & 61.46 & 19.09 & 59.76 & 45.16 & 68.39 \\
    LLaMA-2-7B & 66.08 & 38.91 & 48.78 & 55.92 & 54.80 & 21.66 & 36.80 & 31.34 & 45.82 \\
    \midrule

    \multicolumn{10}{c}{\textit{Different Extension Lengths}} \\
    \midrule
    ET-32k & 98.23 & 80.00 & 86.65 & 93.45 & 78.03 & 44.85 & 85.58 & 47.73 & 78.36 \\
    ET-96k & 97.21 & 72.12 & 86.21 & 91.41 & 86.86 & 41.83 & 83.64 & 45.74 & 76.11 \\
    ET-128k & 99.20 & 73.31 & 53.21 & 76.74 & 84.24 & 39.16 & 83.11 & 46.36 & 72.82 \\
    \midrule

    \multicolumn{10}{c}{\textit{Different Token Quantities}} \\
    \midrule
    ET-2.0B & 95.65 & 79.05 & 85.60 & 85.33 & 70.49 & 42.63 & 79.08 & 47.82 & 75.61 \\
    ET-0.5B & 94.71 & 74.35 & 89.53 & 88.71 & 78.58 & 38.24 & 83.11 & 43.04 & 74.72 \\
    \bottomrule
  \end{tabular}
  }
\end{table}
\begin{table}[!ht]
  \caption{Ablation study on LongBench across different configurations.}
  \label{tab:longbench_ablation_results}
  \centering
  \small
  \renewcommand{\arraystretch}{1.05}
  \resizebox{\textwidth}{!}{
  \begin{tabular}{l cccccc c}
    \toprule
    \multirow{2}{*}{\textbf{Method}} &
    \multicolumn{6}{c}{\textbf{Task}} &
    \multirow{2}{*}{\textbf{Avg.}} \\
    \cmidrule(lr){2-7}
    & \textbf{Single-Doc QA} & \textbf{Multi-Doc QA} & \textbf{Summarization}
    & \textbf{Few-Shot Learning} & \textbf{Synthetic Task} & \textbf{Code Completion} & \\
    \midrule

    \multicolumn{8}{c}{\textit{Main}} \\
    \midrule
    ET & 32.03 & 30.81 & 26.04 & 68.04 & 4.54 & 66.48 & 38.30 \\
    \midrule

    \multicolumn{8}{c}{\textit{Different Base Models}} \\
    \midrule
    Mistral-7B-v0.3 & 34.90 & 32.76 & 23.88 & 65.79 & 3.25 & 59.05 & 37.29 \\
    LLaMA-2-7B & 25.86 & 26.51 & 21.88 & 59.81 & 3.31 & 44.86 & 31.16 \\
    \midrule

    \multicolumn{8}{c}{\textit{Different Extension Lengths}} \\
    \midrule
    ET-128k & 30.18 & 28.41 & 22.68 & 67.21 & 2.27 & 60.41 & 35.68 \\
    ET-96k & 31.51 & 29.80 & 24.90 & 67.89 & 5.43 & 58.49 & 36.88 \\
    ET-32k & 30.30 & 26.23 & 26.19 & 68.42 & 4.50 & 60.39 & 36.45 \\
    \midrule

    \multicolumn{8}{c}{\textit{Different Token Quantities}} \\
    \midrule
    ET-2.0B & 29.03 & 29.61 & 25.37 & 66.29 & 14.00 & 63.44 & 37.86 \\
    ET-0.5B & 28.67 & 27.86 & 25.80 & 67.83 & 12.29 & 65.27 & 37.85 \\
    \bottomrule
  \end{tabular}
  }
\end{table}

\subsection{Ablation Studies}
\label{subsec:ablation}
In this subsection, we conduct extensive ablation studies to evaluate the robustness and scalability of the proposed method across diverse configurations. First, we verify the broad applicability of the proposed approach across different base models, including the architecture of Mistral \citep{jiang2023mistral}. Second, we explore the retention of performance when the context window is expanded to extreme lengths of up to 128K tokens. Third, we investigate the impact of the volume of training data on the consistent scaling of performance. Finally, we examine the sensitivity of the framework to specific variations of the end prompt. The experimental results systematically deconstruct the impact of each factor, demonstrating the stability of the method under varying conditions.

\paragraph{Broad Applicability across Diverse Model Families} 
The proposed method demonstrates broad generalizability across distinct foundation architectures. When integrated into Mistral-7B-v0.3 and LLaMA-2 7B, the framework achieves consistent contextual scaling. The architecture of Mistral reaches average scores of 68.39 on the benchmark of RULER and 37.29 on LongBench, whereas the architecture of LLaMA-2 scores 45.82 and 31.16, respectively. This variance is attributable to the pre-training quality of the underlying models rather than any limitation of the methodology. Thus, the extension technique functions as an effective, model-agnostic approach for long-context processing.

\paragraph{Robust Performance Retention across Extended Context Windows} 
The framework extends the context window to various lengths while preserving high overall performance. On the benchmark of RULER, the model achieves average scores of 78.36 at the length of 32K, 76.11 at 96K, and 72.82 at the extreme context of 128K. This stability is mirrored in LongBench, with scores ranging from 36.45 at 32K to 35.68 at 128K. The minor degradation of performance at extreme lengths is expected given the substantial expansion of the context window, confirming the reliable scalability without catastrophic forgetting.

\paragraph{Consistent Performance Scaling with Increased Training Tokens} 
Scaling the training corpus from 0.5 billion to 2.0 billion tokens yields consistent performance enhancements. The average score increases from 74.72 to 75.61 on the benchmark of RULER, and from 37.85 to 37.86 on LongBench. Specifically, the configuration of 2.0 billion tokens achieves a score of 29.61 in Multi-Doc QA, compared to 27.86 for the setup of 0.5 billion tokens. These results indicate that the method effectively utilizes increased volumes of data to fortify the long-context capabilities.

\paragraph{Consistent Robustness across End Prompt Variations} 
The proposed framework maintains stable performance regardless of the configuration of the end prompt. On the benchmark of LongBench, the average score exhibits a negligible decline from 38.30 (EP\_1) to 37.95 (EP\_3). Similarly, on RULER, the performance ranges from 76.45 (EP\_2) to 74.63 (EP\_3). These minimal fluctuations demonstrate that the extension mechanism is robust and insensitive to specific instructional wording, effectively leveraging generalized termination signals to minimize the need for extensive prompt engineering. A comprehensive description of the specific end prompts is provided in Appendix \ref{subsec:appendix_prompts}.

\begin{figure}[htbp]
  \centering
  \begin{minipage}{0.48\textwidth}
    \centering
    \includegraphics[width=\linewidth]{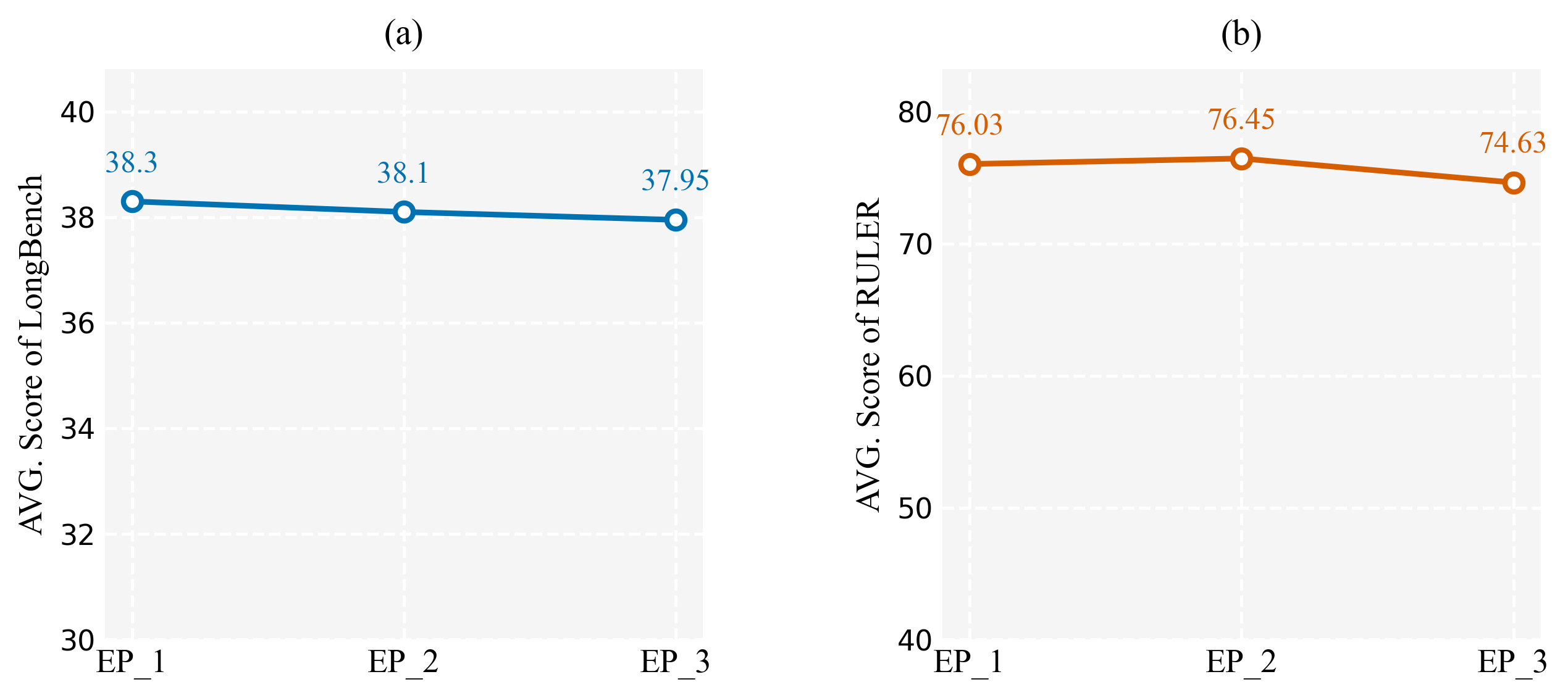}
    \caption{Ablation study on different end prompts.}
    \label{fig:endprompt}
  \end{minipage}\hfill
  \begin{minipage}{0.48\textwidth}
    \centering
    \includegraphics[width=\linewidth]{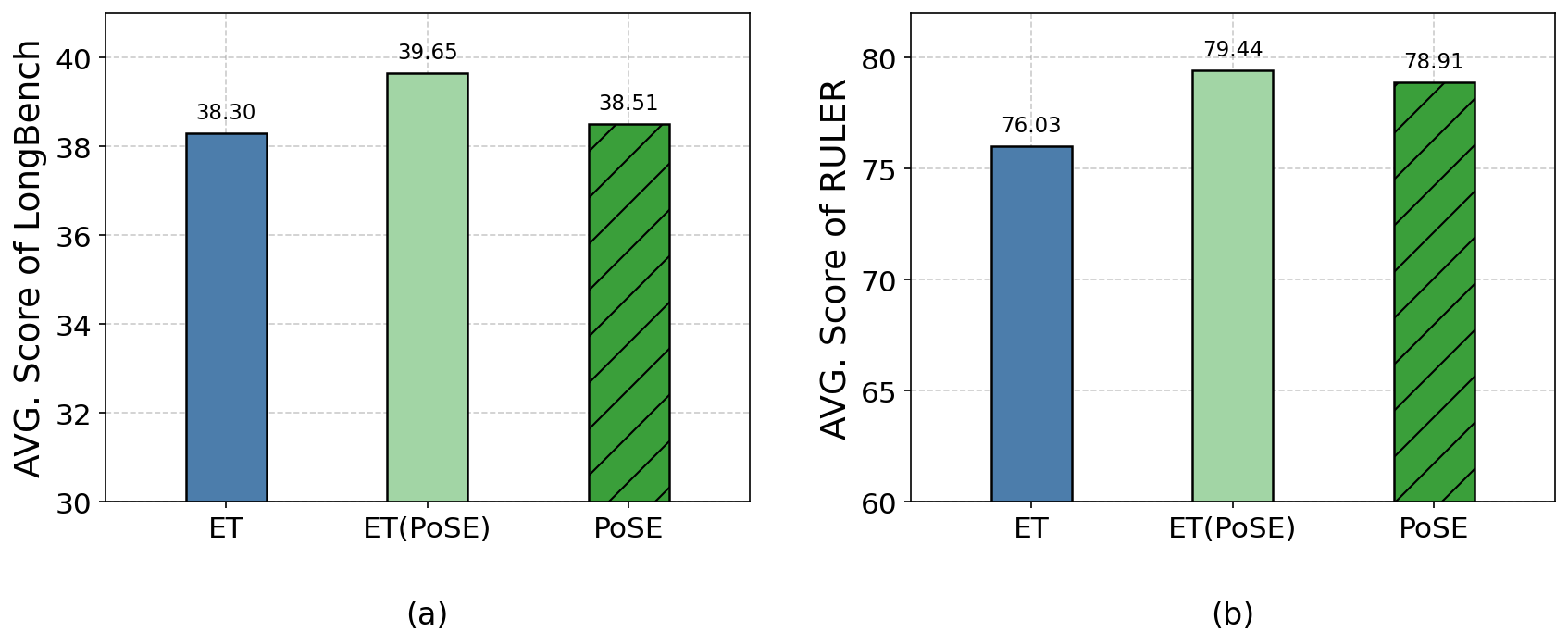}
    \caption{Performance comparison among the standard ET, Positional Skip-Embedding, and ET(PoSE) across the benchmarks of LongBench and RULER.}
    \label{fig:pose}
  \end{minipage}
\end{figure}
\subsection{Structural Analysis}
\label{subsec:structural_analysis}
To evaluate compatibility, the proposed approach is compared against Positional Skip-Embedding, a method that extends the context window by partitioning inputs into chunks and manipulating position indices to simulate extended lengths. We integrate this training methodology with the proposed framework, denoting the hybrid configuration as ET(PoSE). The performances of the standard ET, Positional Skip-Embedding, and the integrated ET(PoSE) are systematically evaluated on LongBench and RULER.
As shown in Figure \ref{fig:pose}, the proposed method demonstrates strong compatibility with Positional Skip-Embedding. On LongBench, the standard ET achieves an average score of 38.30, which is comparable to the 38.51 obtained by Positional Skip-Embedding. On RULER, ET scores 76.03, competitive with the 78.91 achieved by the baseline. Crucially, the hybrid ET(PoSE) configuration yields the highest performance, reaching peak scores of 39.65 on LongBench and 79.44 on RULER. These findings confirm that leveraging the chunking strategy of Positional Skip-Embedding within the ET framework significantly enhances the overall performance.

\subsection{Evaluation and Recovery of Short-Text Capabilities}

\label{subsec:short_text}
This subsection investigates the impact of context extension on the short-text capabilities using the LLaMA-3 8B architecture. The models are evaluated on standard benchmarks, including GSM8K \citep{cobbe2021training}, HumanEval \citep{chen2021evaluating}, MMLU \citep{hendrycks2020measuring}, and HellaSwag \citep{zellers2019hellaswag}. To mitigate performance degradation, a supervised fine-tuning phase is applied. Table \ref{tab:sft_results_highlighted} presents the recovered capabilities. The proposed approach, denoted as sft\_ET(PoSE), achieves the highest average score of 53.56, surpassing full fine-tuning (50.98), LongLoRA (48.64), and LCEG (49.74). Without the integration of Positional Skip-Embedding, the standard ET method still secures a strong average of 52.41. While standalone Positional Skip-Embedding achieves 52.32, the performance discrepancy highlights the effects of the differing training mechanisms. Standard Positional Skip-Embedding partitions the actual training data, risking the disruption of semantic continuity. Conversely, the proposed method partitions only the end prompts, preserving the structural and semantic integrity of the text and ensuring superior performance on short-text tasks.

\begin{table}[htbp]
\caption{Results on short-text evaluation tasks after supervised fine-tuning.}
\label{tab:sft_results_highlighted}
\centering
\begin{tabular}{c|l|cccc c}
\toprule
\multirow{2}{*}{\textbf{Method}} & \multirow{2}{*}{\textbf{Variant}} & \multicolumn{4}{c}{\textbf{Evaluation Task}} & \multirow{2}{*}{\textbf{Avg.}} \\
\cmidrule{3-6}
& & MMLU & GSM8K & HS & HE & \\
\midrule
\multirow{6}{*}{SFT} 
& sft\_ET(PoSE)       & \textbf{56.87} & \textbf{46.10} & \textbf{78.34} & \textbf{32.93} & \textbf{53.56} \\
& sft\_ET             & \underline{56.67} & \underline{44.66} & 77.51 & 30.83 & \underline{52.41} \\
& sft\_LongLoRA       & 55.07 & 38.82 & 77.49 & 23.17 & 48.64 \\
& sft\_LCEG           & 54.76 & 42.23 & 77.56 & 24.39 & 49.74 \\
& sft\_PoSE           & 56.62 & 43.59 & \underline{77.95} & \underline{31.10} & 52.32 \\
& sft\_Full FT        & 55.40 & 42.15 & 77.11 & 29.27 & 50.98 \\
\bottomrule
\end{tabular}
\end{table}

\section{Conclusion}
\label{sec:conclusion}

We presented EndPrompt, an efficient approach for extending the context window of large language models without training on full-length sequences. The method preserves the original short context as an intact segment and appends a brief terminal prompt whose positional indices are placed near the target context boundary. This construction exposes the model to both local and long-range relative distances while avoiding the semantic disruption introduced by chunk-based simulation. Our analysis explains this behavior through RoPE and position interpolation: RoPE represents relative distance through a shared spectral basis, PI suppresses high-frequency positional variation, and shared Transformer parameters couple local and terminal long-distance supervision within the same attention function. Empirically, EndPrompt extends LLaMA-family models from 8K to 64K context and achieves strong results on RULER and LongBench, including a 76.03 average RULER score that exceeds the reported LCEG, LongLoRA, and full-length fine-tuning baselines. These findings suggest that reliable long-context adaptation can be induced from structured sparse positional supervision rather than dense full-length training.

The current formulation relies on an explicit terminal segment. Our future work would study how terminal anchoring can be combined with streaming, hierarchical, or multimodal long-context architectures. More broadly, EndPrompt points to a practical direction for context-window extension -- designing informative positional supervision while keeping the physical training sequence short.

\bibliographystyle{plain}
\bibliography{references}

\newpage
\appendix
\section{Technical Appendices}
\subsection{Experiment details}
We adopt Meta-Llama-3-8B as our base model and employ Position Interpolation (PI) to extend its context window. During data preprocessing, the maximum sequence length of the training corpus is 8K. We perform full-parameter fine-tuning for one epoch on a single computing node equipped with 8 NVIDIA A800 (80GB) GPUs. To optimize memory consumption and training efficiency, all training procedures utilize BF16 mixed precision, integrated with DeepSpeed ZeRO Stage-3, FlashAttention, and gradient checkpointing. For the optimization hyperparameters, we set the peak learning rate to $2 \times 10^{-5}$ and utilize a constant learning rate scheduler with a 20-step linear warmup.

\subsection{Evaluation Tasks}
\label{subsec:tasks}

To thoroughly assess the performance of the models, the experiments incorporate a diverse array of datasets across multiple evaluation frameworks. LongBench \citep{bai2024longbench} provides a comprehensive multi-task evaluation encompassing datasets for single-document and multi-document question answering, document summarization, few-shot learning, synthetic tasks, and code completion. Additionally, RULER \citep{hsieh2024ruler} is utilized to conduct a fine-grained analysis of the effective context length. This benchmark evaluates the information retrieval capabilities of the models through specific dataset configurations, including multi-needle-in-a-haystack tasks, variable tracking, common word extraction, frequent word extraction, and synthetic question answering. These datasets systematically test the extraction of target information embedded within extensive background texts across varying sizes of the context window. Finally, the evaluation of short-text capabilities employs GSM8K \citep{cobbe2021training} for mathematical reasoning, HumanEval \citep{chen2021evaluating} for code generation, MMLU \citep{hendrycks2020measuring} for massive multi-task language understanding, and HellaSwag \citep{zellers2019hellaswag} for commonsense reasoning.

\section{Supplementary Experiment}
\subsection{Detailed End Prompts}
\label{subsec:appendix_prompts}

This section details the specific end prompts evaluated in the robustness analysis. The first configuration, denoted as EP\_1, utilizes the explicit phrase ``This is the end of text, please pay attention here''. The second configuration, EP\_2, employs the native special termination token of the LLaMA-3 architecture, represented as \texttt{<|eot\_id|>}. Finally, the third configuration, EP\_3, consists of the minimal string ``End.''. 
\label{subsec:efficiency}

\subsection{Training Efficacy}
\label{subsec:training_efficacy}
\begin{figure}[htbp]
  \centering
  \includegraphics[width=\linewidth]{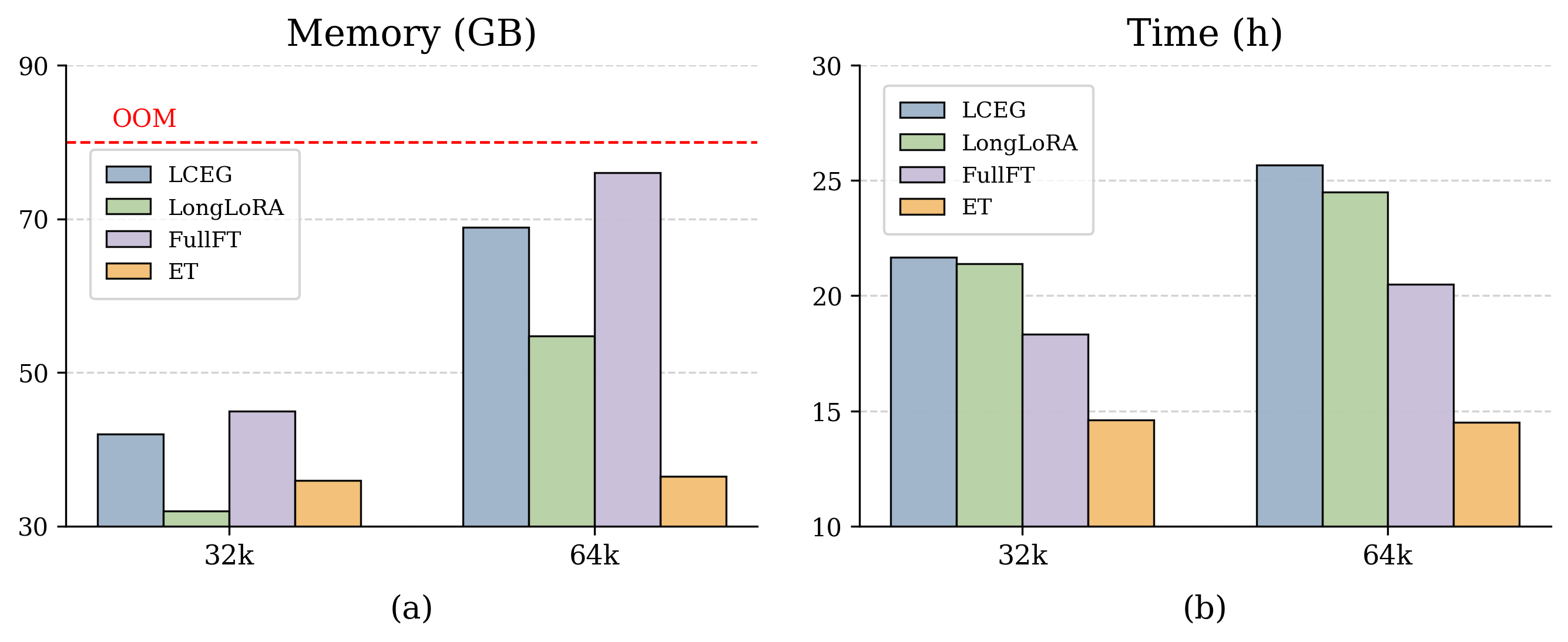}
  \caption{Comparison of memory footprint and time consumption across different methods.}
  \label{fig:memory_time}
\end{figure}

This subsection analyzes the training efficiency by evaluating the memory footprint and the temporal cost. As illustrated in Figure \ref{fig:memory_time}, the ET method overcomes the traditional space-time trade-off, achieving strict dominance over the baseline methodologies at the context lengths of 32K and 64K. At 64K, ET requires only 36.52 GB of memory, representing a 52\% reduction compared to the 76.00 GB required by full-length fine-tuning. Concurrently, ET achieves acceleration ratios of 1.41$\times$, 1.69$\times$, and 1.77$\times$ over full-length fine-tuning, LongLoRA, and LCEG, respectively. This dual advantage of reduced memory utilization and increased computational speed validates the efficiency of the method in scaling context windows without prohibitive overhead.

\subsection{Complete Table\ref{tab:ruler_results}}

\begin{table}[ht]
  \caption{Results on the RULER synthetic benchmark (4K).}
  \label{tab:ruler_results_4k}
  \centering
  \small
  \setlength{\tabcolsep}{15pt} 
  \begin{tabular}{l | c c c c}
    \toprule
    \multirow{2}{*}{\textbf{Task}} & \multicolumn{4}{c}{\textbf{Method}} \\
    \cmidrule(lr){2-5}
    & \textbf{ET} & \textbf{LCEG} & \textbf{LongLoRA} & \textbf{Full FT} \\
    \midrule
    Niah\_S1 & 100.00 & 100.00 & 100.00 & 96.40 \\
    Niah\_S2 & 100.00 & 98.40  & 98.40  & 99.40 \\
    Niah\_S3 & 99.40  & 71.00  & 83.40  & 71.20 \\
    Niah\_M1 & 97.00  & 97.40  & 98.80  & 98.00 \\
    Niah\_M2 & 99.20  & 93.60  & 92.80  & 99.40 \\
    Niah\_M3 & 96.80  & 72.80  & 80.80  & 83.80 \\
    Niah\_MV & 97.20  & 88.90  & 92.40  & 75.40 \\
    Niah\_MQ & 99.15  & 97.45  & 97.20  & 84.50 \\
    Vt       & 99.48  & 95.96  & 93.76  & 93.20 \\
    Cwe      & 91.82  & 88.60  & 83.36  & 82.50 \\
    Fwe      & 90.07  & 62.40  & 61.13  & 76.90 \\
    Qa\_1    & 64.60  & 57.80  & 57.20  & 66.40 \\
    Qa\_2    & 47.20  & 47.00  & 46.00  & 51.80 \\
    \midrule
    Avg.     & \textbf{90.92} & 82.41 & 83.48 & 82.99 \\
    \bottomrule
  \end{tabular}
\end{table}

\begin{table}[ht]
  \caption{Results on the RULER synthetic benchmark (8K).}
  \label{tab:ruler_results_8k}
  \centering
  \small
  \setlength{\tabcolsep}{15pt} 
  \begin{tabular}{l | c c c c}
    \toprule
    \multirow{2}{*}{\textbf{Task}} & \multicolumn{4}{c}{\textbf{Method}} \\
    \cmidrule(lr){2-5}
    & \textbf{ET} & \textbf{LCEG} & \textbf{LongLoRA} & \textbf{Full FT} \\
    \midrule
    Niah\_S1 & 100.00 & 100.00 & 100.00 & 98.00 \\
    Niah\_S2 & 100.00 & 100.00 & 100.00 & 99.00 \\
    Niah\_S3 & 98.80  & 89.20  & 93.40  & 72.60 \\
    Niah\_M1 & 98.80  & 96.20  & 97.40  & 97.60 \\
    Niah\_M2 & 99.20  & 75.40  & 83.20  & 98.60 \\
    Niah\_M3 & 92.00  & 59.20  & 63.20  & 78.00 \\
    Niah\_MV & 96.25  & 93.00  & 94.60  & 87.80 \\
    Niah\_MQ & 96.85  & 96.50  & 97.30  & 87.50 \\
    Vt       & 99.20  & 85.20  & 78.96  & 85.00 \\
    Cwe      & 65.50  & 68.66  & 65.00  & 55.80 \\
    Fwe      & 80.67  & 67.13  & 66.13  & 57.80 \\
    Qa\_1    & 59.40  & 53.80  & 53.40  & 57.80 \\
    Qa\_2    & 45.20  & 46.40  & 46.00  & 50.60 \\
    \midrule
    Avg.     & \textbf{87.07} & 79.28 & 79.89 & 78.93 \\
    \bottomrule
  \end{tabular}
\end{table}

\begin{table}[ht]
  \caption{Results on the RULER synthetic benchmark (16K).}
  \label{tab:ruler_results_16k}
  \centering
  \small
  \setlength{\tabcolsep}{15pt} 
  \begin{tabular}{l | c c c c}
    \toprule
    \multirow{2}{*}{\textbf{Task}} & \multicolumn{4}{c}{\textbf{Method}} \\
    \cmidrule(lr){2-5}
    & \textbf{ET} & \textbf{LCEG} & \textbf{LongLoRA} & \textbf{Full FT} \\
    \midrule
    Niah\_S1 & 100.00 & 100.00 & 100.00 & 98.00 \\
    Niah\_S2 & 100.00 & 100.00 & 100.00 & 96.40 \\
    Niah\_S3 & 97.20  & 91.60  & 93.80  & 79.20 \\
    Niah\_M1 & 95.40  & 95.20  & 96.60  & 95.60 \\
    Niah\_M2 & 95.00  & 63.80  & 71.00  & 95.80 \\
    Niah\_M3 & 67.00  & 34.60  & 48.00  & 62.60 \\
    Niah\_MV & 95.30  & 89.45  & 92.75  & 54.60 \\
    Niah\_MQ & 92.40  & 93.85  & 94.10  & 45.80 \\
    Vt       & 94.52  & 79.88  & 72.96  & 74.20 \\
    Cwe      & 49.76  & 61.94  & 61.22  & 33.30 \\
    Fwe      & 90.00  & 71.07  & 65.53  & 67.20 \\
    Qa\_1    & 61.00  & 53.60  & 53.20  & 58.60 \\
    Qa\_2    & 45.00  & 41.80  & 42.60  & 47.80 \\
    \midrule
    Avg.     & \textbf{83.28} & 75.14 & 76.29 & 69.93 \\
    \bottomrule
  \end{tabular}
\end{table}

\begin{table}[ht]
  \caption{Results on the RULER synthetic benchmark (32K).}
  \label{tab:ruler_results_32k}
  \centering
  \small
  \setlength{\tabcolsep}{15pt} 
  \begin{tabular}{l | c c c c}
    \toprule
    \multirow{2}{*}{\textbf{Task}} & \multicolumn{4}{c}{\textbf{Method}} \\
    \cmidrule(lr){2-5}
    & \textbf{ET} & \textbf{LCEG} & \textbf{LongLoRA} & \textbf{Full FT} \\
    \midrule
    Niah\_S1 & 100.00 & 100.00 & 100.00 & 98.00 \\
    Niah\_S2 & 100.00 & 99.80  & 100.00 & 99.20 \\
    Niah\_S3 & 98.00  & 84.60  & 87.80  & 78.40 \\
    Niah\_M1 & 92.20  & 98.60  & 98.60  & 97.40 \\
    Niah\_M2 & 77.80  & 79.80  & 77.00  & 88.80 \\
    Niah\_M3 & 53.00  & 35.60  & 40.00  & 38.60 \\
    Niah\_MV & 93.75  & 74.85  & 79.60  & 65.30 \\
    Niah\_MQ & 91.10  & 83.65  & 83.45  & 64.90 \\
    Vt       & 93.36  & 66.96  & 56.96  & 55.90 \\
    Cwe      & 6.00   & 41.98  & 39.32  & 14.10 \\
    Fwe      & 88.13  & 61.27  & 55.60  & 44.70 \\
    Qa\_1    & 44.40  & 52.60  & 52.20  & 52.80 \\
    Qa\_2    & 40.20  & 45.20  & 43.40  & 42.20 \\
    \midrule
    Avg.     & \textbf{75.23} & 71.15 & 70.30 & 64.64 \\
    \bottomrule
  \end{tabular}
\end{table}

\newpage
\section{Related Work}
\label{sec:related_work}

\paragraph{Position-based Context Extension}
Large language models predominantly employ Rotary Position Embedding (RoPE) \citep{su2024roformer} to encode positional information. To extend the context length beyond the pretraining limit, Position Interpolation (PI) \citep{chen2023extending} rescales the positional indices to fit within the original domain. Further improvements, such as NTK-aware scaling \citep{peng2023ntk} and YaRN \citep{peng2024yarn}, modify the rotary frequencies based on the neural tangent kernel theory to preserve high-frequency local information while interpolating low-frequency long-range information. While effective, these methods still typically require fine-tuning on long text sequences to achieve optimal performance and alignment.

\paragraph{Efficient Long-Context Adaptation}
Training on full-length sequences incurs quadratic computational complexity. To mitigate this, LongLoRA \citep{chen2024longlora} introduces shifted sparse attention, enabling efficient fine-tuning without significantly modifying the core architecture. Alternatively, simulated long-context training methods like PoSE \citep{zhu2024pose} manipulate positional indices across disjoint chunks of text to simulate long distances. Other orthogonal approaches include RingAttention \citep{liu2024ringattention} for distributed sequence processing across multiple devices, and Activation Beacons \citep{zhang2024soaring} that compress context into condensed representations. In contrast to chunking methods that may break semantic continuity, our approach maintains the intact short context and utilizes an explicit end prompt to anchor long-range positional indices efficiently.

\paragraph{Long-Context Evaluation}
Evaluating extended context capabilities requires rigorous and standardized benchmarks. LongBench \citep{bai2024longbench} and L-Eval \citep{an2024eval} offer diverse multi-task suites encompassing summarization, question answering, and code analysis. ZeroSCROLLS \citep{shaham2023zeroscrolls} focuses specifically on zero-shot long text understanding for summarization tasks. More recently, RULER \citep{hsieh2024ruler} provides a configurable synthetic evaluation framework to test the effective context limits of LLMs by escalating the complexity of needle-in-a-haystack tasks with multi-key and multi-value retrieval scenarios.

\clearpage

\end{document}